\documentclass[11pt,a4paper]{article}
\usepackage[hyperindex,breaklinks]{hyperref}
\usepackage[hyperref]{acl2020}

\usepackage{acl2020}
\usepackage{times}
\usepackage{latexsym}

\usepackage{microtype}
\usepackage{booktabs}
\usepackage{multirow}
\usepackage{array}
\usepackage{diagbox}
\usepackage{paralist}
\aclfinalcopy 

\title{Question and Answer Test-Train Overlap in Open-Domain Question Answering Datasets}

\author{
  Patrick Lewis${}^\dagger{}^\ddagger$, Pontus Stenetorp${}^\ddagger{}$, Sebastian Riedel${}^\dagger{}^\ddagger$\vspace{7pt}\\
  $^\dagger$Facebook AI Research; $^\ddagger$University College London\vspace{2pt}\\
  \texttt{plewis@fb.com}\\
}
\date{}

\begin{document}
\maketitle
\begin{abstract}
Ideally Open-Domain Question Answering models should exhibit a number of competencies, ranging from simply memorizing questions seen at training time, to answering novel question formulations with answers seen during training, to generalizing to completely novel questions with novel answers.
However, single aggregated test set scores do not show the full picture of what capabilities  models truly have. 
In this work, we perform a detailed study of the test sets of three popular open-domain benchmark datasets with respect to these competencies.
We find that 60-70\% of test-time answers are also present somewhere in the training sets.
We also find that 30\% of test-set questions have a near-duplicate paraphrase in their corresponding training sets.
Using these findings, we evaluate a variety of popular open-domain models to obtain greater insight into what extent they can actually generalize, and what drives their overall performance.
We find that all models perform dramatically worse on questions that cannot be memorized from training sets, with a mean absolute performance difference of 63\% between repeated and non-repeated data. Finally we  show that simple nearest-neighbor models 
outperform a BART closed-book QA model, further highlighting the role that training set memorization plays in these benchmarks.
\end{abstract}

\section{Introduction}

Open-domain Question Answering~(ODQA) is a task examining the ability of models to produce answers to natural language factoid questions drawn from an open set of domains. 
ODQA has received significant attention for its potential practical applications, and more recently as a popular method to analyse how well NLP systems can capture and recall factual knowledge.
This interest in ODQA as a challenging ``knowledge-intensive'' task has led to a flurry of recent works that have driven test-set performance on standard ODQA datasets to new heights~\cite[][inter alia]{lee_latent_2019,guu_realm_2020,karpukhin_dense_2020,lewis_retrieval-augmented_2020,izacard_leveraging_2020}.
However, a deeper understanding of what kinds of questions our models can answer well has been less forthcoming.
Whilst there have been several works examining other kinds of QA datasets~\cite{manjunatha_explicit_2018,kaushik_how_2018, sugawara_what_2018,sugawara_assessing_2020}, we know comparatively little about how the questions and answers are distributed in these ODQA benchmarks, making it hard to understand and contextualize the  results we are observing.

In this work, we address these issues via an analysis of the test sets of three popular ODQA datasets, namely WebQuestions~\cite{berant_semantic_2013}, TriviaQA~\cite{joshi_triviaqa:_2017} and Open Natural Questions~\cite{kwiatkowski_natural_2019,lee_latent_2019}. 
We identify three classes of question that a trained ODQA system should be able to answer, in increasing order of difficulty:
\begin{inparaenum}[1)]
\item the most basic behaviour is to be able to reliably recall the answer to a question that the model has seen at training time. 
\item a model should be able to answer novel questions at test time and choose an answer from the set of answers it has seen during training.
\item a strong system should be able to answer novel questions which have answers which are not contained in the training data.
\end{inparaenum}
It is not clear to what extent our current ODQA datasets measure each of these three behaviours. To address this, we stratify the test sets of these datasets.  
Firstly, we split the test data by whether answers in the test set also appear somewhere in the training sets.
We find that  58-71\% of test answers also occur somewhere in the training data, demonstrating that the majority of the test data does not probe for answer generalization.

Secondly, we annotate 1000 question, answer pairs from each test set for repeated questions in their respective training sets. We find that a surprisingly high 28-34\% have paraphrased questions in the training data, the vast majority of which are  near-duplicates differing by one or two words. This result implies that 30\% of the test set of these datasets only  probe for how well models can simply memorize question answer pairs seen at training.

Equipped with these insights, we compute the performance of several recently proposed ODQA models on our test subsets. We test both \emph{Open-book} approaches, which leverage retrieval from a large corpus of documents and \emph{Closed-book} approaches, which focus on training large parametric models with no external knowledge source~\cite{roberts_how_2020}. We find that test data with train-overlapping data contribute the bulk of the overall performance of all the models studied. 

These issues seem to be more acute for closed-book models. Strikingly, we find that a closed-book BART-based model~\cite{lewis_bart_2019} is incapable of producing answers not observed at training time, and achieves very low scores on non-overlapping questions, suggesting this model is only  capable of memorizing question, answer pairs from training time. With this in mind, we build simple nearest-neighbor models which outperform this BART model, despite having virtually no capacity to generalize beyond training data.

To summarize, we make the following contributions:
\begin{inparaenum}[1)]
\item We provide insights into how answer entities are distributed between dataset splits for ODQA datasets
\item We provide annotated subsets of ODQA test sets indicating whether test-time questions are duplicates of training time questions.\footnote{Our data and evaluation code will be made available at \url{https://github.com/facebookresearch/QA-Overlap}}
\item We evaluate a variety of models on our dataset splits, and derive insights into what kinds of question answering behaviour different models achieve.
\end{inparaenum}

\begin{table}[t]
\footnotesize
  \centering
    \begin{tabular}[t]{lcc}
    \toprule
    Dataset & \multicolumn{1}{ m{0.1\textwidth}}{\centering \% Answer overlap} &\multicolumn{1}{m{0.1\textwidth}}{\centering \% Question overlap}  \\
  \midrule
  Natural Questions & 63.6 & 32.5 \\
  TriviaQA & 71.7& 33.6 \\
  WebQuestions & 57.9 & 27.5 \\

     \bottomrule
    \end{tabular}
\caption{Fractions of open-domain test sets that overlap with their training sets.}
\label{overlap_stats}
\end{table}

\section{Datasets}
In our analysis, we consider three widely used Open-domain QA datasets, WebQuestions~\cite{berant_semantic_2013}, TriviaQA~\cite{joshi_triviaqa:_2017}, and Open Natural Questions, a subset of Natural Questions~\cite{kwiatkowski_natural_2019}  introduced by \citet{lee_latent_2019}. All three datasets consist of factual natural language questions and short multi-token answers, but differ slightly in the style of questions and format of answers. 

\paragraph{WebQuestions} WebQuestions is a dataset of 3,778 train and 2,032 test question, answer pairs. Questions were obtained by mining a search engine, and answers are Freebase entities~\cite{bollacker_freebase_2008} annotated by crowdworkers. The ODQA task consists of predicting the name of the freebase entity. We use the standard train/test splits from \citet{berant_semantic_2013}. We use the development split used in \citet{karpukhin_dense_2020}, which was randomly split from the train set. 

\paragraph{TriviaQA} TriviaQA is a dataset of 78,785 train, 8,837 development and 11,313 test question, answer pairs obtained by scraping trivia websites. Answers consist of wikipedia entities, and any alias for the answer entity is considered a correct answer. We use the open-domain train/test splits, which corresponding to the unfiltered-train and unfiltered-dev reading comprehension splits \cite{lee_latent_2019,min_discrete_2019,min_knowledge_2020,karpukhin_dense_2020}.

\paragraph{Open-Natural Questions} Natural Questions consists of search engine questions with answers annotated as spans in wikipedia articles by crowdworkers. The open-domain version of the dataset consists of question, answer pairs from Natural Questions which have short answer spans less than 6 tokens in length. We use the standard open-domain splits in our experiments, consisting of 79,168 train, 8,757 development and 3,610 question answer pairs.

For all three datasets, the canonical train, development and test splits were obtained by randomly splitting the question, answer pairs, and there are no exact duplicate questions in any dataset. We exclude development data from our overlap analyses, and focus purely on train-test overlap to explicitly assess the effects of training memorization.

\section{Test-Train Overlaps}

We explore two ways of examining the test sets based on overlaps between training and test data. Consider a question, answer pair $(q, a)$ from the test set $D_{\textrm{test}}$
 where the answer consists of at least one answer reference $a = \{s_1 .. s_n\}$. We can consider \emph{answer overlap} where there exists at least one $(q',a') \in D_{\textrm{train}}$ which shares at least one answer reference with $(q,a)$. We can also consider \emph{question overlap}, where there exists some $(q'',a'') \in D_{\textrm{train}}$ where $q''$ is a duplicate of $q$, such that $q$ and $q''$ are paraphrases and have the same answer.

\begin{table*}[t]
\footnotesize
  \centering
    \begin{tabular}[t]{ll|ll|ll}
    \toprule
     \multicolumn{2}{c|}{Open Natural Questions} &  \multicolumn{2}{c|}{TriviaQA} &  \multicolumn{2}{c}{WebQuestions} \\
     Overlapping & Non-overlapping & Overlapping & Non-overlapping & Overlapping & Non-overlapping \\
    \midrule
    Phil Simms&Cloves&David Bowie&Death in the afternoon&Harvard&Queen Victoria\\
Brian Johnson&Matt Monro&Battle of camlann&Clash of the Titans&Alderaan&Brasília\\
8&1,020 -- 1,080 kg&Heligoland&ice-cream sundae&India&Paddington\\
the Indians&Hermann Ebbinghaus&Henry VII&Camshaft&2011&Tom Corbett\\
the 1830s&Matt Flinders&Niagra Falls&Cumberland&Zeus&Gary\\

     \bottomrule
    \end{tabular}
\caption{Randomly sampled overlapping and non-overlapping answers from all three test sets.}
\label{answer_overlap_examples}
\end{table*}

\begin{table*}[t]
\footnotesize
  \centering
    \begin{tabular}[t]{lll}
    \toprule
     Answer & Test Question & Train Question \\
    \midrule
    Jason Marsden & who plays max voice in a goofy movie & who does max voice in a goofy movie\\
    January 23 2018 & when will the 2018 oscar nominations be announced & when are the oscar nominations for 2018 announced \\
    Alan Shearer
 & who has scored more goals in the premier league
 & most goals scored by a premier league player \\
retina & where are the cones in the eye located & where are cone cells located in the eye \\
    francisco pizarro & who led the conquest of the incas in south america
 & conquistador who defeated the incan empire in peru
 \\
     \bottomrule
    \end{tabular}
\caption{Randomly sampled test-train overlapping questions in Open Natural Questions. See Appendix \ref{sec:appendi_addtional_qas} for more examples, including examples from TriviaQA and WebQuestions}
\label{question_overlap_examples}
\end{table*}

\paragraph{Answer Overlap}
Following \citet{rajpurkar_squad:_2016}, we apply answer normalization\footnote{
    Answer normalization consists of lower-casing, stripping punctuation, removing articles and normalizing whitespace}
on answer references before searching for overlapping answer references for all $(q,a)$ pairs in the test set
-- see Table~\ref{overlap_stats}.
We find that 58\% of test $(q,a)$ pairs in WebQuestions have answer overlaps, with 63.6\% and 71.7\% for Natural Questions and TriviaQA respectively.
We would naturally expect TriviaQA to have higher answer overlap as it has more answer references per question on average (13.7 references on average compared to 1.2 for Natural Questions and 2.4 for WebQuestions).
Examples of answer overlaps are shown in Table~\ref{answer_overlap_examples}.

\paragraph{Question-Overlap}
Unlike answer overlap, question overlap cannot be easily computed automatically, as searching for duplicates via rules or paraphrase classifiers may lead to both false positives and negatives.
Thus, we turn to manual annotation to investigate question overlap.
To obtain a representative sample for each dataset, we annotate a random subset of 1,000 $(q,a)$ pairs for each test set.
Annotators are shown a list of up to 50 training questions which have a similar answer reference.%
\footnote{
Training questions are selected for annotation if one of the following is true: they share an answer reference with a test question, a test answer reference is a sub-sequence of a training answer reference, or the other way around (a training reference answer is a sub-sequence of a test answer reference).
If there are more than 50 such questions, the top 50 are chosen by the highest degree of word overlap to the test question.
}
This answer similarity function is designed for high recall to obtain a tight lower bound on question overlap.
If there were no questions with similar answers in the training set, the question was automatically annotated as not overlapping.
Three expert annotators looked through these similar questions and indicated if any were paraphrases of the test question and had the same answer.

The results from the annotation can be seen in Table~\ref{overlap_stats} shows these results and examples of overlapping questions in Table~\ref{question_overlap_examples}.
A sample of 100 2-way annotated examples indicated 93\% agreement, corresponding to a Cohen's Kappa of 0.85~\cite{cohen_coefficient_1960}.
What we observe is a high degree of question overlap, with between 27.5 and 33.6\% of the 1,000 annotated test questions had a duplicate in the training set. It is also common to see several duplicates per test question, with an average of 2.8 duplicate questions per overlapping test question in Natural Questions.

\section{Implications for Modelling}

\begin{table*}[t]
\footnotesize
  \centering
\setlength{\tabcolsep}{2pt}
    \begin{tabular}[t]{ll|cccc|cccc|cccc}
    \toprule
    \multicolumn{2}{c|}{\multirow{2}{*}{Model}} &  \multicolumn{4}{c|}{Open Natural Questions} & \multicolumn{4}{c|}{TriviaQA} & \multicolumn{4}{c}{WebQuestions} \\
       \multicolumn{2}{c|}{} &  \multicolumn{1}{m{0.7cm}}{\centering \scriptsize{Total}} &\multicolumn{1}{m{1cm}}{\centering \scriptsize{Question Overlap}} & \multicolumn{1}{m{1cm}}{\centering \scriptsize{Answer Overlap Only}}&  \multicolumn{1}{m{1cm}|}{\centering \scriptsize{No Overlap}} &  \multicolumn{1}{m{0.7cm}}{\centering \scriptsize{Total}} &\multicolumn{1}{m{1cm}}{\centering \scriptsize{Question Overlap}} & \multicolumn{1}{m{1cm}}{\centering \scriptsize{Answer Overlap Only}}&  \multicolumn{1}{m{1cm}|}{\centering \scriptsize{No Overlap}} & \multicolumn{1}{m{0.7cm}}{\centering \scriptsize{Total}} &\multicolumn{1}{m{1cm}}{\centering \scriptsize{Question Overlap}} & \multicolumn{1}{m{1cm}}{\centering \scriptsize{Answer Overlap Only}}&  \multicolumn{1}{m{1cm}}{\centering \scriptsize{No Overlap}} \\
  \midrule
  \multirow{3}{1cm}{Open book}& RAG                       &44.5 & 70.7 & 34.9 & 24.8 & 56.8 & 82.7 & 54.7 & 29.2 & 45.5 & 81.0 & 45.8 & 21.1\\
  & DPR                                                   &41.3 & 69.4 & 34.6 & 19.3 & 57.9 & 80.4 & 59.6 & 31.6 & 42.4 & 74.1 & 39.8 & 22.2\\
  & FID                                                   &51.4 & 71.3 & 48.3 & 34.5 & 67.6 & 87.5 & 66.9 & 42.8 & - & - & - & -\\
  \midrule
  \multirow{2}{1cm}{Closed book}& \scriptsize{T5-11B+SSM} &36.6 & 77.2 & 22.2 & 9.4 & - & - & - & - & 44.7 & 82.1 & 44.5 & 22.0\\
  & BART                                                  &26.5 & 67.6 & 10.2 & 0.8 & 26.7 & 67.3 & 16.3 & 0.8 & 27.4 & 71.5 & 20.7 & 1.6\\
  \midrule
  \multirow{2}{1.1cm}{Nearest Neighbor}& Dense            &26.7 & 69.4 & 7.0 & 0.0 & 28.9 & 81.5 & 11.2 & 0.0 & 26.4 & 78.8 & 17.1 & 0.0\\
  & TF-IDF                                                &22.2 & 56.8 & 4.1 & 0.0 & 23.5 & 68.8 & 5.1 & 0.0 & 19.4 & 63.9 & 8.7 & 0.0\\
    \bottomrule
    \end{tabular}
    \caption{Exact Match scores for several recent models on our dataset splits. The ``Total'' column is the overall performance on the dataset. ``Question Overlap'' refers to the test subset with train-test question overlap, and probes for simple question memorization. ``Answer Overlap Only'' refers to the test subset without train-test question overlap, but with train-test answer overlap, which probes for answer classification.  ``No overlap'' refers to the test subset with no train-test answer overlap and probes for QA generalization}
\label{main_results_table}
\end{table*}

Given our findings from above, we turn our attention to how well ODQA models perform with respect to train-test set overlap. Earlier, we identified three classes of answering behaviors: 1) questions that can be memorized at training time, 2) novel questions that can be answered with answers memorized at training time, 3) novel questions with novel answers. We refer to these behaviours as \emph{Question memorization}, \emph{Answer classification} and \emph{QA generalization} respectively.

\paragraph{Question memorization} To perform well on the question overlap subset, a model would only need to be able to memorize $(q,a)$ pairs at training time, then recognize which training question matches a test-time question.
The reasoning required ranges from trivial duplicate detection for very similar questions such as ``who played pink in pink floyd the wall" and ``who played pink in the movie the wall", to more challenging inference problems for more subtle duplicates such as ``On which island in the North Sea did both St Aidan and St Cuthbert live?'' and ``irish born missionary saint aidan founded a monastery in 653 on which english island which is also the name of a 1970s uk folk-rock band?''. A manual annotation of 100 question-overlap pairs indicated that 81\% were simple duplicates differing by one or two words, 14\% required some paraphrasing recognition capability, and 5\% required more sophisticated natural language understanding. To measure performance on question memorization, we build a test subset comprised of $(q,a)$ pairs which have question overlap to the training set.

\paragraph{Answer Classification} In order to tackle the answer-overlap question, a multi-class classifier over training set answers would be sufficient, as answers never appear at test time that don't appear at training time. We build a test subset of $(q,a)$ pairs which have answer overlap, but do not have question overlap. Question-overlap pairs are excluded to isolate performance on answer classification, since question-overlap questions are significantly easier to answer, and would inflate scores.

\paragraph{QA Generalization} In this regime, models cannot rely on memorizing their training data. To measure performance on this most challenging split, we build a test subset of $(q,a)$ pairs which do not have answer overlap with the training set.
We further note that we expect higher frequency answers, such as countries, integers and public figures would naturally be expected to appear less often in this test subset. As such, models that perform well on the head of the answer distribution may struggle to perform well in this setting, despite being able to perform some generalization at test time.

In the following, we briefly describe the models included in our analysis. For published models, we obtain test set predictions directly from the authors.

\subsection{Open-Book Models}
Open-book Models are ODQA models which first retrieve relevant documents from Wikipedia and then either extract or generate answers conditioned on those documents. 
We consider the Dense Passage Retrieval~(DPR) model~\cite{karpukhin_dense_2020}, a pipeline model which retrieves documents based on dense embeddings, before feeding them into a conventional reader-reranker which extracts spans of text as answers. 
We also include Retrieval-Augmented Generation~\cite{lewis_retrieval-augmented_2020}, a recent model that jointly learns to retrieve and generate answers in seq2seq framework, based on dense retrieval and BART~\cite{lewis_bart_2019}. 
Finally we include the state-of-the-art Fusion-in-Decoder (FID)~\cite{izacard_leveraging_2020}, a pipeline model based on T5-large \cite{raffel_exploring_2020} which retrieves 100 documents and fuses them so that the decoder can attend to all documents at once. We not include FID results on WebQuestions as the authors did not use it in their original work. 

\subsection{Closed-Book Models}
Closed-book models store the knowledge required to answer their questions entirely within the parameters of the model itself, rather than in an external corpus. Typically these models consist of seq2seq transformer models which are directly fine-tuned on $(q,a)$ pairs. In our analysis, we train a BART-large closed-book QA model, which is trained with questions as input and generates $(q,a)$ pairs as output. Checkpoints are selected by Exact Match score on a development set. We also include a much more powerful T5-11B model from \citet{roberts_how_2020}. We use the T5-11B model which has been pretrained with a special ``Salient Span Masking'' objective~\cite{guu_realm_2020}, designed to improve downstream ODQA performance. The T5-11B model was trained on both train and development portions of the data, and thus has seen $\sim$10\% more training data than other models. As we did not include development data in our overlap analysis, a small amount of unaccounted-for overlap is possible for this model. 
We do not include TriviaQA results for the T5 model since this model was trained using a different TriviaQA data splitting scheme.

\subsection{Nearest-Neighbor Models}
Given that there are high levels of train-test overlaps in these datasets, we also experiment with some simple nearest-neighbor models. Here, we simply retrieve a $(q,a)$ pair from the training set based on question similarity to the test question, and return its answer. We experiment with two models, one using TF-IDF and the other using the dot product similarity of question embeddings from the DPR retriever. These models cannot generalize to non-overlapping answers, and have limited capacity to answer non-overlapping questions. However, these models are attractive from the perspective of model size and efficiency. There has recently been a push towards more space and memory-efficient QA systems.\footnote{Such as the EfficientQA competition at Neurips 2020 \url{https://efficientqa.github.io/}} Lightweight retrievers coupled with a database of carefully selected $(q,a)$ pairs would represent a very space-efficient solution compared to open-book models which must retrieve from large textual corpora, or closed-book models with large parameter counts.

\subsection{Results}

Table~\ref{main_results_table} shows our results. In this section we unpack some findings.

\paragraph{Question Memorization} Earlier, we found that $\sim$30\% of test set questions overlap with the training set. The ``Question overlap" columns in Table~\ref{main_results_table} shows performance on Question Memorization. Comparing this column  with the total performance column shows that all models perform significantly higher on memorizable questions. This finding is not surprising, but it is worth highlighting that a significant proportion of overall performance is driven by question memorization. This effect is most pronounced for closed book models. The T5-11B performs especially well for question memorization on both Natural Questions and WebQuestions. This suggests that its very large capacity, coupled with more powerful question understanding may allow it to store, recognise and recall training questions more effectively than other models.

\paragraph{Answer Classification} The ``Answer overlap only" column in Table~\ref{main_results_table} shows performance on answer classification. Answer classification has a large drop in performance compared to question memorization, dropping by an average of 45\% Exact Match score. Open-book models handle this setting better than closed book models. The BART model in particular struggles here, only managing 10.2\% accuracy on this set. 

\paragraph{QA Generalization} The ``No overlap" column in Table~\ref{main_results_table} shows performance on QA generalization.  All models suffer significant performance degradation on QA generalization, highlighting the shortcomings of the overall performance metric. For example, we may expect the FID state-of-the model to answer half of Natural Questions-style questions correctly, but once we have accounted for repeated questions and answers, it can only answer about one third of questions correctly. This difference is even more pronounced for other models, with an average absolute drop of 25\% with respect to overall performance.

\paragraph{Nearest-Neighbor Models} The bottom two rows of Table~\ref{main_results_table} show the results of our nearest-neighbor models. The TF-IDF model, despite being completely untrained, is able to answer about 20\% of test questions correctly, purely by retrieving questions from the training sets. More interestingly, the dense retrieval model outperforms the BART open-domain QA model on Natural Questions and TriviaQA. Furthermore, the dense nearest  neighbor model also outperforms the significantly more complex DPR open-book model on TriviaQA and WebQuestions on the question overlap subset. These models have limitations, but represent very space and memory efficient solutions. Our dense nearest neighbour model consists of a single BERT-base checkpoint and outperforms a BART-large model, and could be compressed using quantization and distillation techniques~\cite{sanh_distilbert_2020,jiao_tinybert_2019}. The TF-IDF model is even smaller and could be implemented extremely efficiently with negligible memory footprint.

\section{Related Work}

The widespread adoption of deep learning in the last few years has been accompanied with an increase in dataset sizes and construction methodologies. Examining what kinds of behaviours are learnt by models has received attention in natural langauge understanding tasks, such as the GLUE benchmark~\cite{wang_glue_2018}, which includes a diagnostic test set probing for different reasoning types.
Various works have also performed critical and careful analysis of question answering systems and datasets. \citet{chen_thorough_2016} closely examine the difficulty of the CNN-DM dataset~\cite{hermann_teaching_2015}, \citet{sugawara_assessing_2020} perform an analysis of machine comprehension dataset difficulty,
\citet{kaushik_how_2018} analyse the difficulty of various machine reading datasets, and \citet{manjunatha_explicit_2018} show that visual question answering models memorize common question-answer relationships present in training data. \citet{fevry_entities_2020} perform an analysis of various closed-book models' TriviaQA predictions, based on entity mentions. \citet{kwiatkowski_natural_2019} note that the machine reading Natural Questions dataset has substantial train-test overlap of wikipedia titles, and provide some baselines for ``long-answer'' QA.
Closest to our work, \citet{verga_facts_2020} observe similar answer overlap in knowledge-base QA, and explore results on non-overlapping subsets.

\section{Conclusion}
In this work, we performed a novel analysis of popular open-domain question answering datasets. We found that 60\% of test set answers overlap with the training set and, more surprisingly, 30\% of test set questions have at least one duplicate in the train set. Following these observations, we contextualize the performance of seven ODQA models, stratifying by different amounts of training set overlap, gaining an insight into to what extent these models generalize or simply memorize their training data. It is clear that performance on these datasets cannot be properly understood by overall QA accuracy and suggest that in future, a greater emphasis should be placed on more behaviour-driven evaluation, rather than pursuing single-number overall accuracy figures.

\section*{Acknowledgments}
The authors would like to thank thank Nicola De Cao, Tom Kwiatkowski, Michael Collins, Kenton Lee, Adam Roberts, Colin Raffel, Scott Yih, Sewon Min, Gautier Izacard and Vladimir Karpuhkin for helpful discussions and providing test set prediction files for analysis.

\clearpage
\bibliography{patrick.bib,additional.bib}
\bibliographystyle{acl_natbib}

\appendix

\section{Appendices}
\label{sec:appendix}
\subsection{Additional Question Overlap Examples}
\label{sec:appendi_addtional_qas}
Tables \ref{question_overlap_extra_examples}, \ref{question_overlap_triviaqa} and \ref{question_overlap_webquestions} give more question overlap examples for the three datasets.
\begin{table*}[t]
\footnotesize
\centering
    \addtolength{\leftskip}{-3cm} 
    \addtolength{\rightskip}{-3cm}
    \begin{tabular}[t]{p{3.1cm}p{6.9cm}p{7.5cm}}
    \toprule
     Answer & Test Question & Train Question \\
    \midrule
    Bob Geldof & who played pink in pink floyd the wall & who played pink in the movie the wall\\
    Daren Maxwell Kagasoff & who played ricky in secret life of the american teenager & who played ricky on the secret life of the american teenager \\
Andy & who does april end up with on parks and rec & who does april marry in parks and rec \\
may 5 2017 & when did gaurdians of the galaxy 2 come out & when is guardians of the galaxy vol 2 released \\
norman pritchard & who won the first medal in olympics for india &  who won the first individual olympic medal for india \\
moira kelly & who does the voice of nala in the lion king & who played nala in the lion king movie \\
supreme court & who enforces the charter of rights and freedoms & who has final authority of interpretation of the canadian charter of rights and freedoms \\
554 & most passing yards by nfl qb in a game & what is the nfl record for most passing yards in a single game\\
John Ross & who ran the fastest 40 yard dash in the nfl & who has the fastest 40 yard dash ever \\
international border ib & what is the name of india pakistan border & what is the border name between india and pakistan \\
Andrew Wright & who wrote when a man loves a woman & who wrote song when a man loves a woman \\
new england patriots & who has participated in the most super bowls & what nfl team has been to most super bowls \\

     \bottomrule
    \end{tabular}
\caption{Additional examples of test-train overlapping questions in Open Natural Questions}
\label{question_overlap_extra_examples}
\end{table*}

\begin{table*}[t]
\footnotesize
\centering
    \addtolength{\leftskip}{-3cm} 
    \addtolength{\rightskip}{-3cm}
    \begin{tabular}[t]{p{3.1cm}p{6.9cm}p{7.5cm}}
    \toprule
     Answer & Test Question & Train Question \\
    \midrule
    Picasso & Who painted "Boy With a Pipe" which, in May 2004, was sold for a record price of \$104 million? & painted in 1905, the painting garcon a la pipe was a famous painting by which famous artist who died in 1973? \\
    Wensum & On what river is the city of Norwich & the english city of norwich lies on which river?\\
    Mantle & Comprising around two-thirds of the Earth's mass , what is found between the core of the Earth and its crust? & what do we call the layer of the earth between its crust and its core?\\
    Live and Let Die & In which James Bond film does actress Jane Seymour play Solitaire?
 & jane seymour played the character "solitaire" in which bond film? \\
Esau &
Who, in the Bible, was the eldest son of Isaac? &
in the bible, who was the first born of isaac? 
\\
Alanis Morrisette &
Who made the 1995 album 'Jagged Little Pill' which sold 33 million copies? &
who released the 1995 hit album "jagged little pill"?
\\
Excalibur &
In British legend, what is the name of King Arthur’s sword? &
what was the name of king arthur's sword?
\\
Humidity &
What is measured by a Hygrometer? &
what does a hygrometer measure?
\\
A Storm&
On the Beaufort scale what is defined as force 11?&
what is force 11 (eleven) on the beaufort scale?
\\
Jeremy Irons&
Actress Sinead Cusack is married to which 'Oscar' winning actor?&
which actor is the husband of sinead cusack?
\\
Sir Cloudesley Shovell &
Who was the British Admiral who died in 1707 when four of his ships were wrecked in the Scilly Isles? &
in 1707 a fleet of navy ships was wrecked off the scilly islands. who was the commander who lost his life in the disaster?\\
Tony Hart &
Which famous individual created the 'Blue Peter' sailing ship logo? &
which artist designed the logo for uk television children’s show ‘blue peter’?\\
\bottomrule
    \end{tabular}
\caption{Examples of test-train overlapping questions in TriviaQA}
\label{question_overlap_triviaqa}
\end{table*}

\begin{table*}[t]
\footnotesize
\centering
    \addtolength{\leftskip}{-3cm} 
    \addtolength{\rightskip}{-3cm}
    \begin{tabular}[t]{p{3.1cm}p{6.9cm}p{7.5cm}}
    \toprule
     Answer & Test Question & Train Question \\
    \midrule

costa rica &
where is isthmus of panama located on the map?&
where is isthmus of panama located?\\

1986 world series&
when's the last time the mets won the world series?&
when did the mets win the pennant?\\

abbottabad&
where was bin laden found and killed?&
what country was osama bin laden killed in?\\

believer&
what other movies has ryan gosling been in?&
what movies does ryan gosling star in?\\

sculpture &
what type of art did leonardo da vinci make?&
what kind of art did leonardo da vinci produce?\\

origin of species&
what book did charles darwin wrote in 1859?&
what was the name of the book that charles darwin wrote?\\

morehouse college&
what college did martin luther king jr go to?&
where did dr. martin luther king jr. go to school?\\

communist state&
what type of government did soviet union have?&
what type of government does the former soviet union have?\\

turkish lira&
what money to take to turkey?&
what currency to take to side turkey?\\

spanish language&
what is the most common language spoken in argentina?&
what is language in argentina?\\

opera OR classical music&
what music period did beethoven live in?&
what music did beethoven composed?\\

harry s truman&
who was president after franklin d. roosevelt?&
who became president when roosevelt died in office?\\

\bottomrule
    \end{tabular}
\caption{Examples of test-train overlapping questions in WebQuestions}
\label{question_overlap_webquestions}
\end{table*}

\end{document}